\pdfoutput=1

\documentclass[11pt]{article}

\usepackage{acl}

\usepackage{times}
\usepackage{latexsym}

\usepackage[T1]{fontenc}

\usepackage[utf8]{inputenc}

\usepackage{microtype}

\usepackage{algorithm}
\usepackage{enumerate}
\usepackage{enumitem}
\usepackage{multirow}
\usepackage{graphicx}
\usepackage{array}
\usepackage{booktabs}
\usepackage{tikz}
\usetikzlibrary{patterns}
\usepackage{pgf-pie}
\usepackage{colortbl}
\usepackage{array}   
\usepackage{fdsymbol}
\usepackage{inconsolata}
\usepackage{verbatim}
\usepackage{stmaryrd}
\usepackage{trimclip}
\usepackage[tikz]{bclogo}
\usepackage{makecell}
\usepackage{pifont}

\usepackage{pifont}

\definecolor{applegreen}{rgb}{0.56, 0.8, 0.25}
\definecolor{mypurple}{rgb}{0.62,0.24,0.81}
\usepackage{algpseudocode}
\newcommand{\ie}{{\em i.e.,}}
\newcommand{\eg}{{\em e.g.,}}
\newcommand{\Ni}{({\em i})~}
\newcommand{\Nii}{({\em ii})~}
\newcommand{\Niii}{({\em iii})~}

\newcommand{\Na}{({\em a})~}
\newcommand{\Nb}{({\em b})~}

\usepackage[nameinlink]{cleveref}
\crefformat{section}{\S#2#1#3} 
\crefname{algorithm}{Alg.}{Algs.}
\crefformat{subsection}{\S#2#1#3}
\Crefname{equation}{Eq.}{Eqs.}
\Crefname{figure}{Fig.}{Figs.}
\definecolor{msftBlack}{RGB}{0,0,0}
\newcommand{\finding}[1]{
	\begin{bclogo}[couleur= msftBlack!10, epBord=1, arrondi=0.1, logo=\bclampe, marge=2, ombre=true, blur, couleurBord=msftBlack!20, tailleOndu=3, sousTitre={\em #1}]{} 
	\end{bclogo}
}

\algdef{SE}[DOWHILE]{Do}{doWhile}{\algorithmicdo}[1]{\algorithmicwhile\ #1}%
\title{Relevant or Random: Can LLMs Truly Perform Analogical Reasoning?}

\author{Chengwei Qin\textsuperscript{\ding{70}}$^\dagger$\thanks{\; Equal contribution, order decided by coin flip.}, Wenhan Xia$^\clubsuit$\footnotemark[1], Tan Wang$^\dagger$\footnotemark[1], Fangkai Jiao$^\dagger$, Yuchen Hu$^\dagger$, \\\textbf{Bosheng Ding$^\dagger$}, \textbf{Ruirui Chen$^\vardiamondsuit$}, \textbf{Shafiq Joty$^\dagger$$^\spadesuit$}\\
\textsuperscript{\ding{70}}The Hong Kong University of Science and Technology (Guangzhou) $^\clubsuit$Princeton University\\
$^\dagger$Nanyang Technological University $^\spadesuit$Salesforce Research\\
$^\vardiamondsuit$Institute of High Performance Computing (IHPC),\\
Agency for Science, Technology and Research (A*STAR), Singapore
}

\begin{document}
\maketitle

\begin{abstract}

{Analogical reasoning is a unique ability of humans to address unfamiliar challenges by transferring strategies from relevant past experiences.} One key finding in {psychology} is that compared with irrelevant past experiences, recalling \emph{relevant} ones can help humans \emph{better} handle new tasks. Coincidentally, the NLP community has also recently found that self-generating relevant examples in the context can help large language models (LLMs) better solve a given problem than hand-crafted prompts. 
However, it is yet not clear whether relevance is the key factor eliciting such capability, \ie\ can LLMs benefit more from self-generated relevant examples than irrelevant ones? In this work, we systematically explore whether LLMs can truly perform analogical reasoning on a diverse set of reasoning tasks. With extensive experiments and analysis, we show that self-generated random examples can surprisingly achieve comparable or even better performance on \emph{certain} tasks, \eg\ 4$\%$ performance boost on GSM8K with random biological examples. We find that the accuracy of self-generated examples is the key factor and subsequently design two novel methods with improved performance and significantly reduced inference costs.
Overall, we aim to advance a deeper understanding of LLM analogical reasoning and hope this work stimulates further research in the design of self-generated contexts.

\end{abstract}

\section{Introduction} \label{sec:intro}

\begin{figure}[t]
    \centering
    \includegraphics[width=0.44\textwidth]{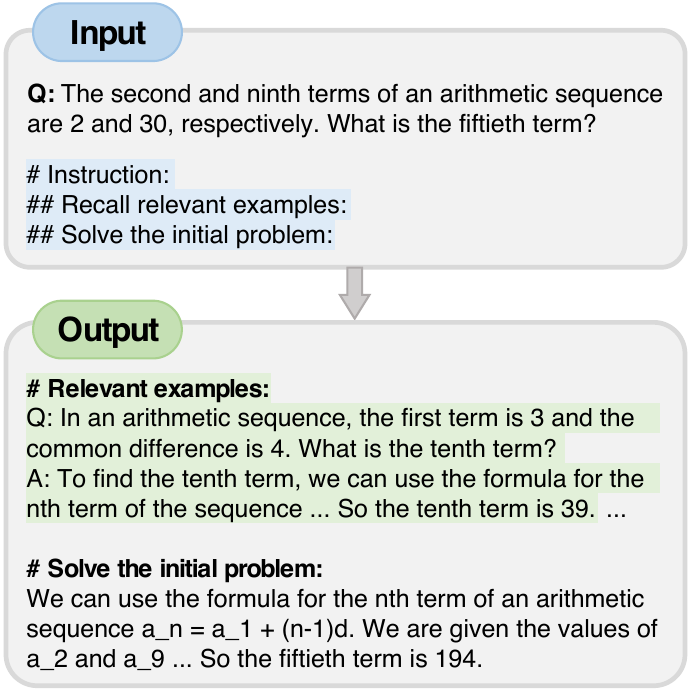}
    \caption{Illustration of LLM analogical reasoning in \citet{yasunaga2024large}. LLMs are prompted to self-generate relevant examples as context before solving the new problem.}
    \label{fig:analogicalprompt}
\end{figure}

A hallmark of human intelligence is that they can solve novel problems by drawing analogy from relevant past experiences, a concept known as \emph{analogical reasoning} in cognitive science \citep{vosniadou1989similarity}. As indicated by the name, recalling previously acquired \emph{relevant} experiences can facilitate humans to \emph{better} tackle new tasks, whereas irrelevant ones are rarely beneficial and can even be distracting \cite{Gentner12Analogical}. For instance, when faced with a novel math problem about determinants (\eg\ calculating the value of a given fourth-order determinant), humans can resolve this by reflecting upon the methodology employed to ascertain the value of a third-order determinant, whereas biological knowledge (\eg\ how the human body regulates its temperature)  can generally be considered irrelevant.

With the recent advancements in scaling up model size and data, LLMs have demonstrated impressive zero-shot and few-shot performance across various reasoning tasks, especially, through advanced prompting methods like chain-of-thought (CoT) \citep{cot_wei}. Compared to common approaches such as zero or few-shot CoT \citep{zhou2022least,kojima2022large,zhang2022automatic}, \citet{yasunaga2024large} introduce LLM analogical reasoning, \ie\ LLMs self-generate examples relevant to the query as context to better solve new problems; see \Cref{fig:analogicalprompt} for an example. 
However, it remains unclear whether relevance is the key to eliciting such capability in LLMs. While several studies explore the influence of the relevance of demonstrations in in-context learning (ICL) and CoT \citep{liu-etal-2022-makes,kim2022self,lyu-etal-2023-z,chen-etal-2023-self,yang2023auto,wang-etal-2023-towards,alkhamissi-etal-2023-opt,yasunaga2024large,luo2024let}, none of them investigate whether self-generated relevant examples consistently outperform irrelevant ones in LLM analogical reasoning.

In this paper, to systematically assess the capability of LLMs to perform analogical reasoning, we conduct a series of ablation experiments on a variety of reasoning tasks including problems from GSM8K \citep{cobbe2021training}, MATH \citep{hendrycks2021measuring}, and BIG-Bench Hard (BBH) \citep{suzgun2022challenging}. Furthermore, we evaluate the generalizability of our findings to other reasoning tasks, \eg\ GPQA \citep{rein2024gpqa}, in Section~\ref{sec:general_diff_tasks}. With extensive experiments, we aim to address the following two research questions:

\begin{itemize}[leftmargin=*,topsep=4pt,itemsep=2pt,parsep=2pt]
    \item \textbf{Q1.} Are self-generated \emph{relevant} examples more beneficial to LLMs than \emph{random} ones? 
    \item \textbf{Q2.} {If not, what is the pivotal factor for LLMs' performance in analogical reasoning?}
\end{itemize}

To answer these questions, we empirically analyze the analogical reasoning abilities of GPT-3.5 (turbo), GPT-4o-mini, the Llama series \citep{touvron2023llama}, and Qwen 2.5 \citep{yang2024qwen2} models. Surprisingly, experimental results show that prompting LLMs to self-generate random examples can achieve comparable or even better performance on \emph{certain} tasks which is not in line with the key claim of analogical reasoning in \citet{Gentner12Analogical}, indicating that LLMs \emph{cannot always} perform analogical reasoning. As for Q2, we point out through controlled experiments that the key factor is \textit{the accuracy of self-generated examples}. Informed by these findings, we design two approaches that can outperform existing methods with significantly reduced inference costs. Specifically, we ask LLMs to randomly generate a few problems and manually verify their correctness, then use this fixed set of problems as in-context learning demonstrations for all test samples. Consistent observations across different model types and scales consolidate the conclusions. We summarize the major contributions of our work below:

\begin{itemize}[leftmargin=*,topsep=4pt,itemsep=2pt,parsep=2pt]
    \item {To the best of our knowledge, we, for the first time, extensively assess the ability of LLMs to perform analogical reasoning and explore their counterintuitive behavior on certain tasks}.
    \item With extensive experiments and analysis, we demonstrate the effectiveness and limitations of different types of self-generated contexts. 
    \item Building on the findings, we propose two novel ICL-based approaches that improve performance while significantly reducing inference costs.
\end{itemize}

\begin{figure*}[ht]
  \centering
\includegraphics[width=0.94\textwidth]{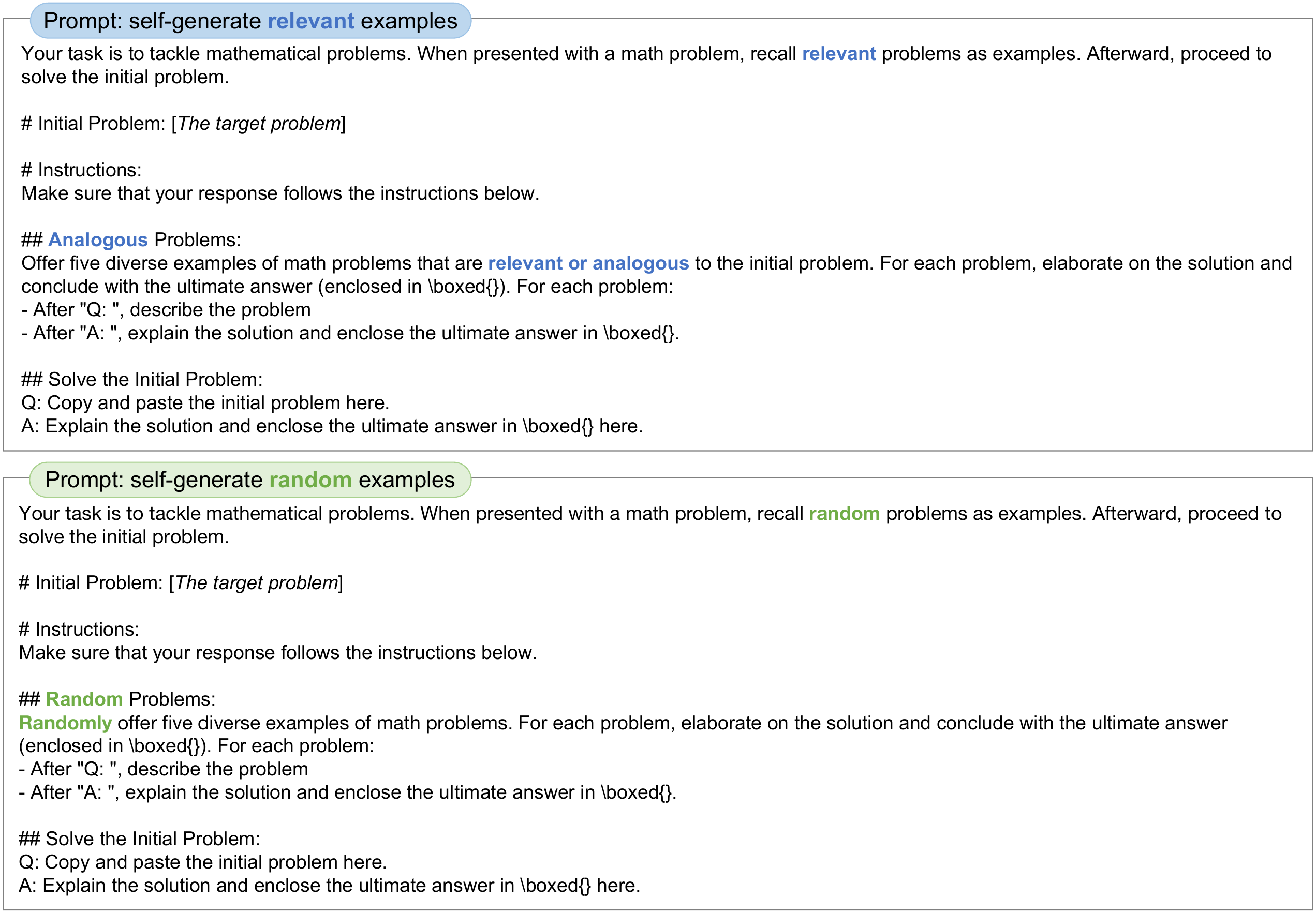}
  \caption{Example prompts for GSM8K (mathematical reasoning). \emph{\textbf{Top}}: The original prompt used in \citet{yasunaga2024large} for self-generating \emph{relevant} math problems. \emph{\textbf{Bottom}}: The prompt designed for self-generating \emph{random} math problems. We mark the differences between these two prompts in \textcolor{blue}{blue} and \textcolor{applegreen}{green} respectively.}
  \label{fig:example_prompt}
\end{figure*}

\section{Related Work}

{This work mainly explores whether LLMs can truly perform analogical reasoning. In light of this, we review two lines of research that form the basis of this work: chain-of-thought prompting and LLM analogical reasoning.}

\subsection{Chain-of-Thought Prompting}

{Chain-of-thought (CoT) prompting induces LLMs to generate intermediate reasoning steps before generating the final answer \citep{cot_wei}, greatly improving the reasoning capabilities of LLMs. Typical CoT prompting approaches include few-shot CoT \citep{cot_wei,zhou2022least,cot_wei_sc,li2022advance,wang2022rationale}, taking several labeled demonstrations of the reasoning process, and zero-shot CoT, comprising only instructions like ``Let's think step by step'' \citep{kojima2022large,zelikman2022star,zhang2022automatic}. Other ongoing research on CoT has also explored \Ni optimizing the demonstration selection~\citep{fu2022complexity,li-qiu-2023-finding,qin-etal-2024-context}, \Nii optimizing the quality of reasoning chains~\citep{khot2022decomposed,chen2022program,shinn2023reflexion,zhao-etal-2023-verify,besta2024graph}, and \Niii CoT in smaller models \citep{magister2022teaching,ho2022large,fu2023specializing,qin2023improving,ranaldi-freitas-2024-aligning,peng2025lmm}.}

\subsection{LLM Analogical Reasoning}

While few-shot CoT can provide more detailed reasoning guidance, it requires labeled examples which can be unavailable for a new task. To tackle this problem, \citet{yasunaga2024large} propose analogical prompting to guide LLMs to self-generate relevant exemplars as few-shot demonstrations, which is similar to analogical reasoning, \ie\ humans can address new problems by drawing analogy from relevant past experience  \citep{vosniadou1989similarity,holyoak2012analogy}. LBS3 \citep{luo2024let} explores curriculum learning which can better reflect human learning habits. In this work, we step forward to explore the intrinsic principle of LLM analogical reasoning. Specifically, we aim to investigate whether LLMs can authentically exhibit such reasoning capabilities and determine the extent to which the relevance of self-generated examples contributes to enhancing this process.

\section{Methodology}

Our analysis is based on the analogical prompting approach outlined in \citet{yasunaga2024large}. Specifically, for a given target problem $x$, analogical prompting introduces instructions like:

\begin{center}
	\parbox{0.43\textwidth}{\# Problem: [$x$]\vspace{0.45em}}
        \parbox{0.43\textwidth}{\# Relevant problems: Recall five \underline{relevant} and diverse problems. For each problem, describe it and explain the solution.\vspace{0.45em}}
        \parbox{0.43\textwidth}{\# Solve the initial problem:}
\end{center}

The goal is to induce LLMs to self-generate \emph{relevant} examples, aiding them to solve the target problem via in-context learning. To ensure better performance and efficiency, several key technical decisions are made in \citet{yasunaga2024large}:

\begin{itemize}[leftmargin=*,topsep=2pt,itemsep=2pt,parsep=0pt]
    \item {The self-generated examples should be relevant and diverse, achieved through a specially designed instruction.}
    \item {Generate relevant problems and the solution to the initial problem in one pass.}
    \item {$3$ to $5$ self-generated examples perform the best.}
\end{itemize}

In this work, we leverage similar prompts\footnote{Since our work aims to comprehensively explore and analyze the intrinsic principle of LLM analogical reasoning proposed in \citet{yasunaga2024large}, we should  follow the original design of the instructions to have a fair comparison and reliable analysis.} to guide LLMs to generate different types of \emph{irrelevant} examples as context; {see \Cref{fig:example_prompt} for example prompts}:

\begin{itemize}[leftmargin=*,topsep=2pt,itemsep=2pt,parsep=0pt]
    \item \emph{$\text{N/A}$}: generate problems that are N/A (not applicable) to the initial problem.
    \item \emph{$\text{Random}_{\text{same}}$}: randomly generate examples of the same problem type (\eg\ math).
    \item \emph{$\text{Random}_{\text{diff}}$}: randomly generate examples of different problem types (\eg\ any type except math).
    \item \emph{$\text{Random}_{\text{bio}}$}: randomly generate biological problems.
\end{itemize}

\citet{yasunaga2024large} demonstrate that self-generating relevant examples can consistently outperform zero-shot CoT and few-shot CoT (hand-crafted examples or retrieved top-$k$ most similar training samples) on different tasks. Therefore, we do not include these two methods in our work. Interested readers can refer to the corresponding results and analysis in \citet{yasunaga2024large}. In addition, we show prompts for different methods on all datasets in \Cref{sec:example_prompt_allmethods}.

\section{Experiment} \label{sec:exp}

\begin{table*}[t]    
    \centering
    \scalebox{0.94}{
    \begin{tabular}{
        l @{\hspace{2em}}
        cccccc
        }
        \toprule
        \textbf{Method} & \makecell{Temporal \\ sequences}  & \makecell{Logical deduction \\ five objects} & \makecell{Reasoning about \\ colored objects} & \makecell{Formal \\ fallacies} &  \makecell{Word \\ sorting} & Average \\
        \midrule
        Relevant & \textbf{60.0} & \textbf{51.2} & \textbf{76.7} & 51.2 & 76.9 & \textbf{63.2} \\
        \midrule
        N/A & 57.5 & 45.3 & 75.5 & \textbf{53.3} & \textbf{77.7} & 61.9 \\
        $\text{Random}_{\text{same}}$ & 53.1 & 48.8 & 73.5 & 52.4 & 74.1 & 60.4  \\
        $\text{Random}_{\text{diff}}$ & 44.3 & 44.8 & 72.4 & 51.2 & 69.2 & 56.4 \\
        $\text{Random}_{\text{bio}}$  & 57.1 & 49.5 & 76.1 & 50.8 & 74.9 & 61.7 \\
        \bottomrule
    \end{tabular}
    }
    \caption{ Accuracy ($\%$) of different methods on five reasoning tasks in BBH. \textbf{Bold} indicates the best results. Self-generated \emph{relevant} examples achieve the best average performance. Detailed results for different seeds are reported in \Cref{sec:detailed_res_diff_seeds}.
    }
    \label{tab:bbh-result}
\end{table*}

\begin{table}[t]
\centering
    \scalebox{0.94}{
    \begin{tabular}{lccc}
    \toprule
    \multirow{2}{*}{\textbf{Method}} & \multicolumn{3}{c}{\textbf{Task}} \\
    \cmidrule(lr){2-4}
    & GSM8K & MATH & Average  \\
    \midrule
    Relevant & 71.5 & 33.3 & 52.4 \\
    \midrule
     N/A & 75.5 & 36.1 & \textbf{55.8} \\
    $\text{Random}_{\text{same}}$ & 75.1 & \textbf{36.3} & 55.7 \\
    $\text{Random}_{\text{diff}}$ & \textbf{76.3} & 34.1 & 55.2 \\
    $\text{Random}_{\text{bio}}$ & 75.3 & 34.6 & 54.9 \\
    \bottomrule
    \end{tabular}
    }
\caption{
\label{tab:math_reasoning}
Accuracy ($\%$) of different methods on two mathematical reasoning tasks. Self-generated \emph{irrelevant} examples are consistently better than \emph{relevant} ones. \Cref{tab:detailed_results_all_seeds_math} in \Cref{sec:detailed_res_diff_seeds} reports detailed results for different seeds. 
}
\end{table}

\subsection{Experimental Setup}

We construct the evaluation suite based on diverse reasoning-intensive tasks, including mathematical reasoning and other reasoning (\eg\ logical and temporal reasoning) tasks:

\begin{itemize}[leftmargin=*,topsep=3pt,itemsep=3pt,parsep=0pt]

\item \textbf{Mathematical reasoning}. We work with two commonly used datasets, GSM8K \citep{cobbe2021training} and MATH \citep{hendrycks2021measuring}. For each dataset, we randomly sample 500 examples from the original test set and run experiments three times with different random seeds (resulting in different test samples). 

\item \textbf{Other reasoning}. Following \citet{yasunaga2024large}, we evaluate five reasoning tasks in BIG-Bench Hard (BBH) \citep{suzgun2022challenging}: temporal sequences (temporal reasoning), logical deduction five objects and reasoning about colored objects (logical reasoning), formal fallacies (deductive reasoning) and word sorting (symbolic reasoning). For each task, we use all test samples for evaluation and run experiments three times with different random seeds.
\end{itemize}

We mainly use GPT-3.5 (gpt-3.5-turbo) as the LLM (see \Cref{sec:results_gpt_4o_mini} for more results with GPT-4o-mini) and obtain all outputs from it with the temperature set to 0. We ask the LLM to self-generate 5 examples for GSM8K, 3 examples for MATH and BBH following \citet{yasunaga2024large}.

\subsection{Main Results} \label{sec:main_res_analogical}

We now address the research questions asked in \Cref{sec:intro} with empirical results. 

\finding{\small \textbf{Q1.} Are self-generated relevant examples more beneficial to LLMs than random ones?   \label{Q1}
}

{The results averaged over all random seeds are reported in \Cref{tab:bbh-result} and \Cref{tab:math_reasoning}; more detailed results for every seed are shown in \Cref{sec:detailed_res_diff_seeds}.}

\paragraph{$\bullet$ Self-generated relevant examples achieve the best average performance on BBH.} 
From the results in \Cref{tab:bbh-result}, we can observe that the superiority of self-generated relevant examples is empirically substantiated on BBH. {Specifically, using relevant examples, denoted by `relevant', outperforms other approaches on temporal and logical reasoning tasks.} While it performs worse than `N/A' on deductive and symbolical reasoning, it can still improve the accuracy by \textbf{1.3}$\%$ on average compared to `N/A'. 

{However, the results on mathematical reasoning tasks are quite counterintuitive as described below}:
\paragraph{$\bullet$ Relevant examples do not guarantee better performance.} Different from BBH, all types of self-generated irrelevant examples consistently outperform relevant ones on both mathematical reasoning datasets, showing that LLMs cannot yet perform analogical reasoning on these tasks. Interestingly, when we use randomly generated biological examples (\eg\, how the process of photosynthesis occurs in plants), they can yield about \textbf{2.5}$\%$ better results on average compared to generating relevant math problems. Besides, `N/A' achieves the best average result as it is the second-best on both datasets.

\begin{table*}[t]    
    \centering
    \scalebox{0.91}{
    \begin{tabular}{
        l @{\hspace{2em}}
        ccccccc
        }
        \toprule
        \textbf{Method} & \makecell{Precalculus}  & \makecell{Intermediate \\ Algebra} & \makecell{Algebra} & \makecell{Prealgebra} &  \makecell{Counting \& \\ Probability} & Geometry  & \makecell{Number \\ Theory}\\
        \midrule
        Relevant & 10.4	& 9.8 &	51.8 &	56.8 &	22.1 &	24.2 &	\textbf{37.0} \\
        \midrule
        N/A & 9.1 &	15.7 &	\textbf{55.5} &	\textbf{61.0} &	\textbf{28.7} &	\textbf{25.8} &	34.2 \\
        $\text{Random}_{\text{same}}$ & 12.3 &	\textbf{17.6} &	54.4 &	60.6 &	25.4 &	\textbf{25.8} &	34.9  \\
        $\text{Random}_{\text{diff}}$ & \textbf{13.0} &	14.1 &	52.7 &	56.8 &	26.2 &	24.2 &	33.6 \\
        $\text{Random}_{\text{bio}}$  & \textbf{13.0} &	12.2 &	53.0 &	59.2 &	\textbf{28.7} &	\textbf{25.8} &	32.2 \\
        \bottomrule
    \end{tabular}
    }
    \caption{{Accuracy ($\%$) across different subjects in the MATH dataset. Self-generated irrelevant examples outperform relevant ones on 6 out of 7 subjects.}
    }
    \label{tab:math-diff-sub}
\end{table*}

\begin{figure}[t]
    \centering
    \includegraphics[width=0.44\textwidth]{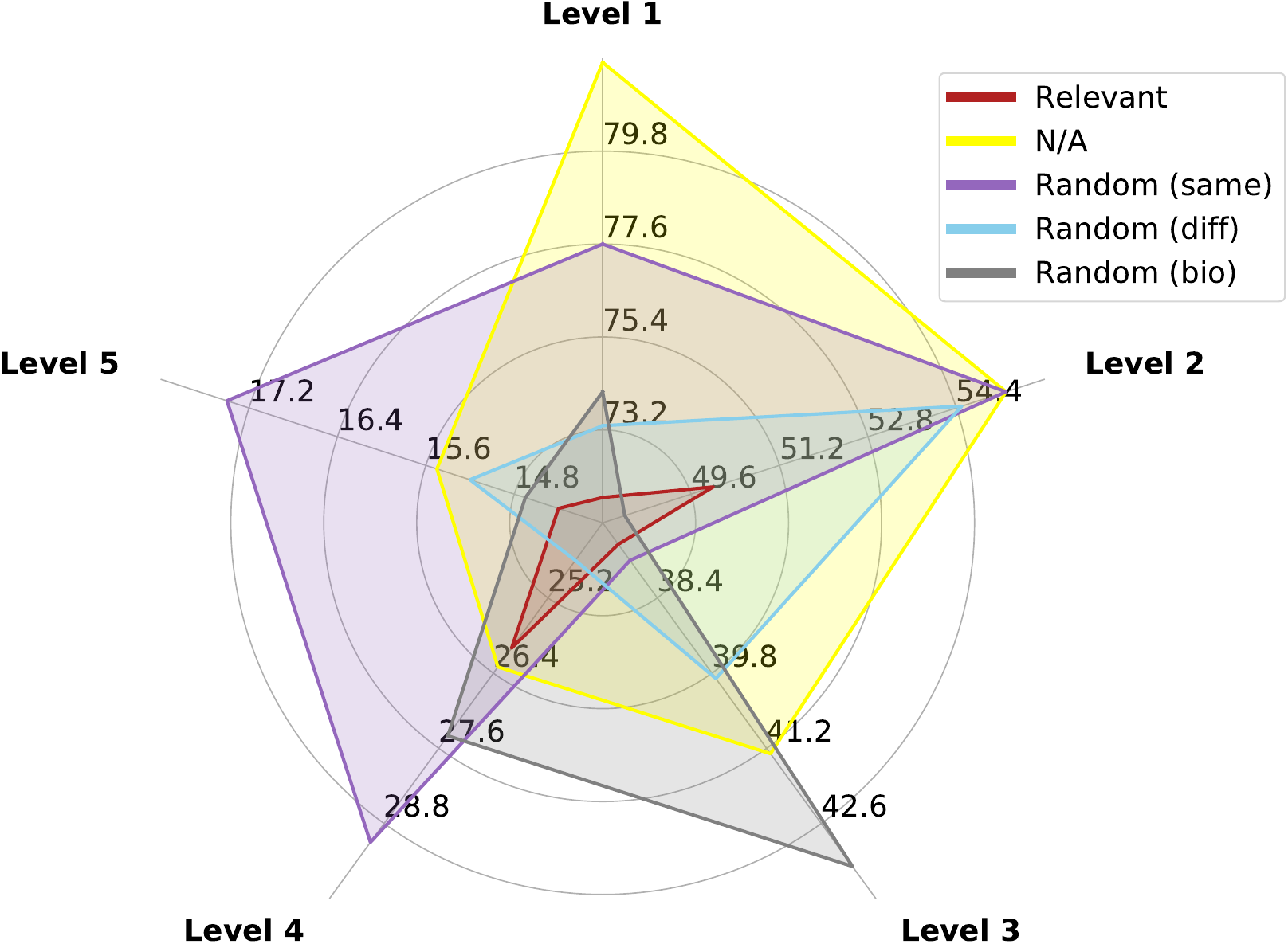}
    \caption{{Comparison of all methods at different difficulty levels on MATH. Level 1 represents the easiest and level 5 is the hardest. `relevant' clearly performs worse than other approaches at all difficulty levels.}}
    \label{fig:diff_level_math}
\end{figure}

{Problems in MATH span various subjects and difficulty levels. To investigate whether the inferior performance of relevant examples on MATH is accidentally caused by certain categories, we further report the accuracy across different subjects and difficulty levels in \Cref{tab:math-diff-sub} and \Cref{fig:diff_level_math}. The consistent performance gap between `relevant' and other methods across different problem categories demonstrates the inherent flaws of relevant examples, indicating that \emph{mathematical reasoning tasks exhibit different analogical reasoning paradigms from BBH.}}

{It might present challenges to prompt LLMs to accurately generate specific types of demonstrations. Therefore, given the unexpected results on mathematical reasoning tasks, one may wonder:}

\finding{\small \textbf{Q1-1.} {Are self-generated examples really relevant or irrelevant to the query?}  \label{Q1_1}
}

\begin{table}[t]    
    \centering
    \scalebox{0.91}{
    \begin{tabular}{
        l @{\hspace{2em}}
        ccc
        }
        \toprule
        \textbf{Method} & GSM8K & MATH & Average\\
        \midrule
        Relevant & \textbf{0.54} &	\textbf{0.41}	& \textbf{0.48} \\
        \midrule
        N/A & 0.19 &	0.28 &	0.24 \\
        $\text{Random}_{\text{same}}$ & 0.30 &	0.20 &	0.25 \\
        $\text{Random}_{\text{diff}}$ & 0.15 &	0.10 &	0.13 \\
        $\text{Random}_{\text{bio}}$  & 0.06 &	0.11 &	0.09 \\
        \midrule
        Oracle & 0.65 &	0.63 & 0.64 \\
        \bottomrule
    \end{tabular}
    }
    \caption{{Average relevance score (semantic similarity) between self-generated examples and the query. `Oracle' stands for the average similarity score between the query and $k$ most similar training samples ($k$ is the number of self-generated examples).}
    }
    \label{tab:example_relevance}
\end{table}

\begin{table}[t]    
    \centering
    \scalebox{0.91}{
    \begin{tabular}{lccc}
        \toprule
          & Relevant  & N/A & $\text{Random}_{\text{same}}$\\
        \midrule
        Accuracy & 62.0 & 72.0 & 86.0  \\
        \bottomrule
    \end{tabular}
    }
    \caption{{Accuracy ($\%$) of self-generated examples on the MATH dataset. The examples generated by `relevant' are less accurate.}
    }
    \label{tab:example_acc_math}
\end{table}

\begin{table*}[t]
\centering
\scalebox{0.82}{\begin{tabular}{ll}
\toprule
\multicolumn{2}{l}{\textbf{Query:} For how many ordered pairs $(A,B)$ where $A$ and $B$ are positive integers is $AAA_7+BBB_7=666_7?$ }   \\ 
\midrule
\multicolumn{1}{l}{Relevant}     & \makecell[l]{In a certain base, the sum of two three-digit numbers is $777$. If the digits of one of the numbers are 
\\reversed, the sum becomes $888$. What is the base of this number system?} \\ \hline
\multicolumn{1}{l}{N/A}    & What is the value of $x$ in the equation $2x + 5 = 10$?  \\ \hline

\multicolumn{1}{l}{$\text{Random}_{\text{same}}$}          & \makecell[l]{In a bag, there are 5 red marbles, 3 blue marbles, and 2 green marbles. If you randomly pick 2 marbles \\ from the bag without replacement, what is the probability that both marbles are red?}  \\ \hline

\multicolumn{1}{l}{$\text{Random}_{\text{diff}}$}     & How do you bake chocolate chip cookies? \\ \hline

\multicolumn{1}{l}{$\text{Random}_{\text{bio}}$}    & How does the process of photosynthesis occur in plants?       \\ \hline
\multicolumn{1}{l}{Oracle}          & Find the number of ordered pairs $(a,b)$ of complex numbers such that $a^3 b^5 = a^7 b^2 = 1$. 
\\ \bottomrule
\end{tabular}}
\caption{{Demonstration examples of different methods on the MATH dataset. The example generated by `relevant' is more related to the query than other examples generated by `N/A' or `random'.}}
\label{table:case-study-example}
\end{table*}

\begin{table*}[t]    
    \centering
    \scalebox{0.94}{
    \begin{tabular}{lcccccc}
    \toprule
        \multirow{2}{*}{\textbf{Variant}} & \multicolumn{3}{c}{\textbf{GSM8K}} & \multicolumn{3}{c}{\textbf{MATH}} \\
        \cmidrule(lr){2-4}  \cmidrule(lr){4-7}
        & Relevant  & N/A & $\text{Random}_{\text{same}}$ & Relevant  & N/A & $\text{Random}_{\text{same}}$ \\
        \midrule
        ICL &71.2 &73.8 &72.0 & 37.0 & 39.8 & 39.2 \\
        GPT4-Calibration &\textbf{75.2}  &\textbf{75.6}  &\textbf{75.6}  & \textbf{44.4}  & \textbf{41.2} & \textbf{40.0} \\
        Random &70.0  &72.0  &68.4  & 36.0 & 38.0 & 37.8\\
        \bottomrule
    \end{tabular}
    }
    \caption{ Accuracy ($\%$) of different variants on GSM8K and MATH. When using GPT4-generated answers (mostly accurate), `GPT4-Calibration' consistently outperforms `ICL' for all methods. In contrast, `random' always performs worse than `ICL'. 
    }
    \label{tab:analysis_reason}
\end{table*}

{To quantitatively measure the relevance between the generated examples and the query, we compute the average cosine similarity between them. Following \citet{zhang2022automatic}, we use the sentence transformer \citep{reimers-gurevych-2019-sentence} to encode all samples. For each method, the reported result is averaged across three seeds (see \Cref{sec:decomposition_similarity} for the decomposition of relevance).} 

As observed from \Cref{tab:example_relevance}, relevant examples are much more semantically similar to the query than irrelevant ones and the relevance score of `relevant' is more biased towards `oracle' rather than `random' or `N/A', demonstrating that \emph{LLMs indeed follow instructions to generate specific types of demonstrations}. Furthermore, we calculate the average similarity score between self-generated relevant examples and queries for BBH (0.46), which is slightly lower than the score of mathematical reasoning tasks (0.48). This result demonstrates that the difference in analogical reasoning performance between BBH (\Cref{tab:bbh-result}) and mathematical reasoning (\Cref{tab:math_reasoning}) is \emph{not} because LLMs can generate more relevant examples for BBH.

{We provide a case study in \Cref{table:case-study-example} to delve deeper into the demonstrations of different methods. As we can notice, the example generated by `relevant' is more related to the query as they both involve the mathematical concept `number bases'. In contrast, examples such as `What is the value of $x$ in the equation $2x + 5 = 10$?' (N/A) or `How do you bake chocolate chip cookies?' ($\text{Random}_{\text{diff}}$) are less relevant to the query. This comparison highlights once again that relevance may not be the key factor for analogical reasoning performance on mathematical reasoning tasks. To understand better the underlying reasons for the counterintuitive results, we then ask the following question:}

\finding{\small \textbf{Q2.} {If relevance is not the key factor, what is more important for the accuracy of analogical reasoning?}  \label{Q2}
}

{Looking back at \Cref{table:case-study-example}, an interesting observation is that the self-generated relevant example appears to be more difficult to solve than the irrelevant ones, regardless of whether they are math problems or not. Consequently, the accuracy of relevant examples may be lower. To verify this, we conduct a pilot experiment on MATH. Specifically, we randomly select 50 samples for different types of generated math problems, \ie\ Relevant, N/A and $\text{Random}_{\text{same}}$, and manually evaluate their accuracy. We exclude other methods as it is difficult to define the `accuracy' of the examples they generate. From the results in \Cref{tab:example_acc_math}, we can observe that while the examples generated by `relevant' are more related to the test query, \emph{they are less accurate}, raising the question whether the performance of different approaches on mathematical reasoning tasks is strongly correlated with the accuracy of self-generated examples.}

\begin{figure*}[t]
  \centering
    \includegraphics[width=0.94\textwidth]{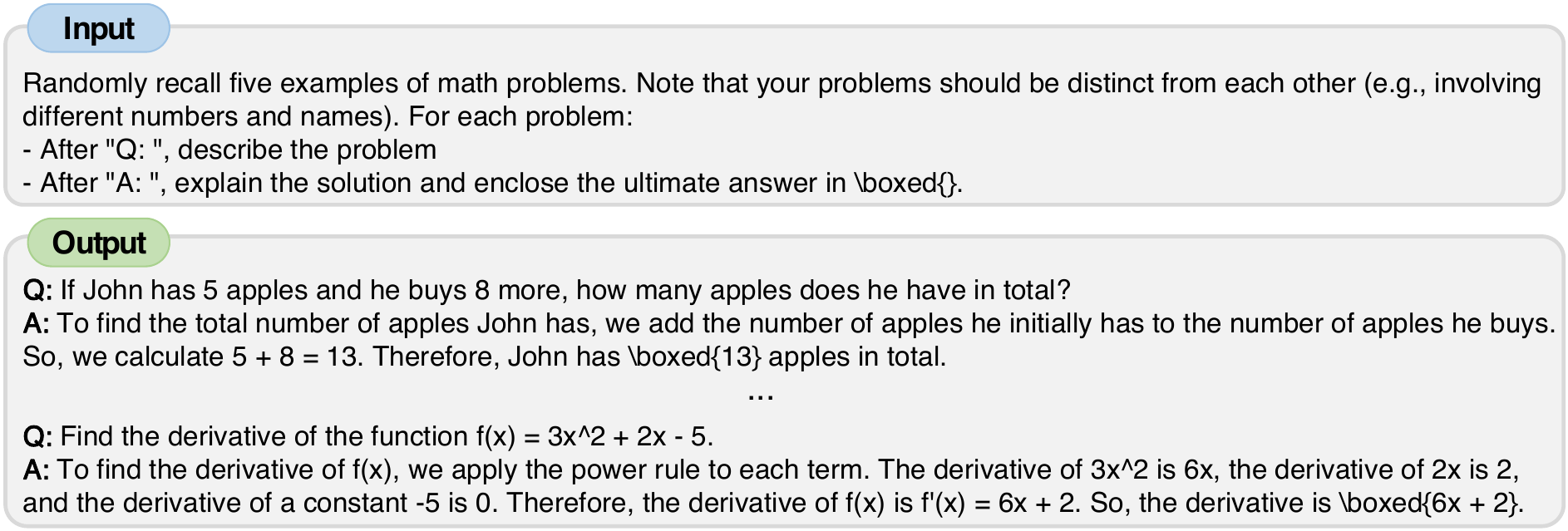}
  \caption{Example prompts and outputs for randomly generating math problems. We manually verify the answers to ensure the correctness of the generated examples.}
  \label{fig:prompt_generating_example}
\end{figure*}

\paragraph{Proxy Approaches} However, as the accuracy of the examples located at the output cannot be directly controlled, we meticulously design a variant called \emph{ICL}, which extracts the generated examples from the model output as in-context learning (ICL) demonstrations and combines them with the query as input to LLMs, as a proxy for the original method.
We also consider the following two variants: \Na \emph{GPT4-Calibration} which replaces the answers of demonstrations in \emph{ICL} with GPT4-generated answers, and \Nb \emph{Random} changes the answers of demonstrations in \emph{ICL} to random numbers. Our manual verification confirmed that GPT4-generated answers were mostly accurate. {We conduct this experiment on GSM8K and MATH with GPT-3.5 as the LLM reasoner.}

\begin{table}[t]
\centering
    \scalebox{0.94}{
    \begin{tabular}{lccc}
    \toprule
    \multirow{2}{*}{\textbf{Method}} & \multicolumn{3}{c}{\textbf{Task}} \\
    \cmidrule(lr){2-4}
    & GSM8K & MATH & Average  \\
    \midrule
    Relevant & 71.5 & 33.3 & 52.4 \\
    N/A & 75.5 & 36.1 & 55.8 \\
    $\text{Random}_{\text{same}}$ & 75.1 & 36.3 & 55.7 \\
    $\text{Random}_{\text{diff}}$ & 76.3 & 34.1 & 55.2 \\
    $\text{Random}_{\text{bio}}$ & 75.3 & 34.6 & 54.9 \\
    \midrule
    $\text{ICL}_{\text{math}}$ & 75.7 & \textbf{36.8} & 56.3 \\
    $\text{ICL}_{\text{bio}}$ & \textbf{77.9} & 34.9 & \textbf{56.4} \\
    \bottomrule
    \end{tabular}
    }
\caption{
\label{tab:comparison_icl}
{Comparison of different methods on two mathematical reasoning tasks.} 
}
\end{table}

From the results of different variants reported in \Cref{tab:analysis_reason}, we can see that increasing the accuracy of generated examples can indeed improve the performance: \emph{GPT4-Calibration} consistently outperforms \emph{ICL} by incorporating more accurate answers. In contrast, \emph{random} always performs the worst among all variants. Therefore, the key factor influencing the performance on mathematical reasoning is \emph{the accuracy of self-generated examples} rather than their relevance. 

It is worthwhile to note that while several papers explore how the correctness of demonstration answers influences in-context learning \citep{min-etal-2022-rethinking,yoo-etal-2022-ground,wei2023larger,pan-etal-2023-context,kossen2024incontext}, our work differs from them in the following aspects: \Ni The examples in our work are generated by LLMs rather than real data from NLP benchmarks, \ie\ randomly sampled from the training set. In addition, there are rationales (CoT) in self-generated examples, which are different from the input-label format of in-context learning investigated in these papers; and \Nii These studies mainly evaluate in-context learning on different classification or multi-choice datasets, \ie\ the output space is a finite set. In contrast, we are evaluating mathematical reasoning tasks, where the output space is infinite.

{Given the above findings, a natural question is}: 

\finding{\small \textbf{Q2-1.} {Can we ask the LLM to randomly generate a few math or biological problems and manually verify their correctness, then use this fixed set of problems as ICL demonstrations for all test queries?} \label{Q2_1}
}

{We refer to these two methods as $\text{ICL}_{\text{math}}$ and $\text{ICL}_{\text{bio}}$, and conduct experiments with them on GSM8K and MATH (see \Cref{fig:prompt_generating_example} for example prompts and outputs for generating math problems). Detailed prompts and outputs for different methods are provided in \Cref{sec:detail_prompt_generating_problems}. Following the original setting, we ask the LLM to randomly generate 5 examples for GSM8K and 3 examples for MATH. As observed from \Cref{tab:comparison_icl}, ensuring the accuracy of self-generated examples does lead to better performance regardless of the problem type. $\text{ICL}_{\text{math}}$ and $\text{ICL}_{\text{bio}}$ achieve similar average performance, once again demonstrating that relevance does not matter (see \Cref{sec:guided_problem_generation} for more analysis on relevance). Moreover, both ICL variants only need to generate examples once, which significantly reduces the inference cost and further demonstrates their superiority.

\begin{table*}[t]    
    \centering
    \scalebox{0.88}{
    \begin{tabular}{lccccccc}
        \toprule
         \textbf{Method} & Relevant  & N/A & $\text{Random}_{\text{same}}$ & $\text{Random}_{\text{diff}}$ &  $\text{Random}_{\text{bio}}$ & $\text{ICL}_{\text{math}}$ &  $\text{ICL}_{\text{bio}}$ \\
        \midrule
        Llama-2-70b-Chat & 45.1 & 51.4 & 50.9 & 54.3 & 47.1 & 55.5 & \textbf{56.1} \\
        Llama-3-8B-Instruct & 69.5 & 72.3 & 72.6 & 74.1 & 73.5 & 75.8 & \textbf{76.8} \\
        Llama-3.1-8B-Instruct & 74.8  &  77.3  &  78.4  &  78.8  &  77.6  & 80.2   &  \textbf{81.0} \\
        Qwen2.5-14B-Instruct & 86.5  &  89.1 & 88.2  & 89.7  & 88.4  & \textbf{91.1}  & 90.6 \\
        \bottomrule
    \end{tabular}
    }
    \caption{Accuracy ($\%$) of different methods on GSM8K using Llama-2-70b-Chat, Llama-3-8B-Instruct, Llama-3.1-8B-Instruct and Qwen2.5-14B-Instruct models. Self-generated relevant examples always perform worse than irrelevant ones and both ICL variants outperform other approaches.
    }
    \label{tab:diff_model_math}
\end{table*}

\begin{table}[t]    
    \centering
    \scalebox{0.88}{
    \begin{tabular}{lccc}
    \toprule
        \multirow{2}{*}{\textbf{Variant}} & \multicolumn{3}{c}{\textbf{Method}} \\
        \cmidrule(lr){2-4}
        & Relevant  & N/A & $\text{Random}_{\text{same}}$  \\
        \midrule
        ICL &56.2 &	58.2 &	58.6 \\
        GPT4-Calibration & \textbf{60.8} &	\textbf{61.0} &	\textbf{60.8}   \\
        Random & 53.2 & 54.0 & 59.6   \\
        \bottomrule
    \end{tabular}
    }
    \caption{ Accuracy ($\%$) of different variants on GSM8K using Llama-2-70b-Chat. `GPT4-Calibration' consistently performs better than `ICL' and `random'. 
    }
    \label{tab:diff_model_reason}
\end{table}

\subsection{Further Analysis} \label{sec:further analysis}

\paragraph{Difference from Previous Work} Apart from the comprehensive analysis, we have designed two novel ICL-based approaches that are completely different from the one in \citet{yasunaga2024large} (Q2-1). The difference lies mainly in the following two aspects: \Ni The key claim in \citet{yasunaga2024large} is that we should guide the model to self-generate relevant examples as context. Motivated by the analysis and findings in our work (Q1 and Q2), our methods focus on ensuring the accuracy of self-generated examples rather than their relevance, which leads to better performance regardless of the problem type. \Nii As we have demonstrated that the relevance of self-generated examples does not matter, there is no need to generate relevant examples for each test query (the original method in \citet{yasunaga2024large}). In contrast, our methods use a fixed set of examples for all test queries, which significantly reduces the inference cost.

\begin{table*}[t]    
    \centering
    \scalebox{0.88}{
    \begin{tabular}{lccccccc}
        \toprule
         \textbf{Dataset} & Relevant  & N/A & $\text{Random}_{\text{same}}$ & $\text{Random}_{\text{diff}}$ &  $\text{Random}_{\text{bio}}$ & $\text{ICL}_{\text{same}}$ &  $\text{ICL}_{\text{bio}}$  \\
        \midrule
         CSQA & 70.8 & 73.4  & 71.2  & 72.9  & 72.6 & \textbf{74.6} & 74.1 \\
          MBPP & 58.2 &  59.8  &  60.6 & 59.6 & 60.2 & \textbf{62.0} & 61.4 \\
        GPQA & 31.6 & 34.4  & 33.7  & 33.1   & 32.6  & 35.8  & \textbf{36.2} \\
        \bottomrule
    \end{tabular}
    }
   \caption{
\label{tab:result_csqa}
Accuracy ($\%$) of different methods on CommonsenseQA, MBPP, and GPQA. `same' in $\text{ICL}_{\text{same}}$ stands for `generating \emph{correct} problems of the \emph{same} type as the dataset'.
}
\end{table*}

\paragraph{Generalization to Open-Source LLMs} Our experiments and analysis so far used GPT-3.5 as the LLM, which is closed-source and gets updated over time. To verify whether the observations and conclusions are consistent across different models and additionally for reproducibility, we extend the experiments to Llama-2-Chat \citep{touvron2023llama}. 
{Specifically, we use vLLM to serve a Llama-2-70b-Chat model for experiments and report the results of different methods/variants on GSM8K in \Cref{tab:diff_model_math} and \Cref{tab:diff_model_reason}.} 
{We can draw similar observations: \Ni self-generated relevant examples underperform all types of irrelevant ones, \Nii `GPT4-Calibration' consistently outperforms the other two variants, and \Niii $\text{ICL}_{\text{math}}$ and  $\text{ICL}_{\text{bio}}$ perform better than other approaches, demonstrating that the conclusions can be generalized to different models.} 

We further conduct experiments with Llama-3-8B-Instruct, Llama-3.1-8B-Instruct \citep{dubey2024llama} and Qwen2.5-14B-Instruct \citep{yang2024qwen2}. The results reported in \Cref{tab:diff_model_math} demonstrate the generalizability of the conclusions across different model types and scales. In addition, since investigating analogical reasoning requires LLMs to self-generate different types of problems, we only experiment with instruction-tuned LLMs to ensure that they can follow the given instructions.

\paragraph{Generalization to Different Tasks} \label{sec:general_diff_tasks} To test the generalizability of our findings beyond the math domain, we further conduct experiments on CommonsenseQA (commonsense reasoning) \citep{talmor-etal-2019-commonsenseqa}, MBPP (code generation) \citep{austin2021program} and GPQA (question answering of very hard questions) \citep{rein2024gpqa}. The comparison between different methods is shown in \Cref{tab:result_csqa}, which demonstrates that our findings can be generalized to different types of tasks.

\paragraph{Comparison Beyond Analogical Reasoning} We consider two widely used methods Self-consistency \citep{wang2023selfconsistency} and Auto-CoT \citep{zhang2023automatic}, and compare our designed approaches with them on GSM8K using Llama-3.1-8B-Instruct. For Self-consistency, we employ 5 decoding paths for majority voting. The results reported in \Cref{tab:comparison_beyond_analogical} demonstrate that our methods can also outperform other baselines beyond analogical reasoning.

In addition, we show the robustness to prompt format, the effect of the number of demonstrations, more analysis on $\text{ICL}_{\text{math}}$ and $\text{ICL}_{\text{bio}}$, the results of repeating problems and explicitly controlling the semantics of generated examples in Appendix \ref{sec:robust_prompt} $\sim$ \ref{sec:explicit_semantic}, respectively.

\begin{table}[t]    
    \centering
    \scalebox{0.72}{
    \begin{tabular}{ccccc}
        \toprule
         Relevant  & Self-consistency & Auto-CoT & $\text{ICL}_{\text{math}}$ &  $\text{ICL}_{\text{bio}}$ \\
        \midrule
         74.8 & 77.6  & 75.9  & 80.2 & \textbf{81.0}  \\
        \bottomrule
    \end{tabular}
    }
   \caption{
\label{tab:comparison_beyond_analogical}
Comparison between our designed methods and baselines beyond analogical reasoning.
}
\end{table}

\section{Conclusion}

In this work, we have systematically assessed the capability of LLMs to perform analogical reasoning. We have identified key research questions and empirically analyzed a representative set of LLMs on a diverse collection of reasoning tasks. Extensive experimental results and analysis show that LLMs \emph{cannot always} perform analogical reasoning and the key influencing factor is the accuracy of self-generated examples rather than their relevance. Given these findings, we have designed two ICL-based approaches with better performance and significantly reduced inference costs.
In the future, we would like to investigate additional analogical prompting methods to generate more accurate examples.

\section*{Limitations}

This work has several limitations. First, due to the inference cost of ChatGPT, we conduct experiments on subsets of the test data for mathematical reasoning tasks. Besides, we include 6 datasets requiring different reasoning capabilities in this work. A further improvement could be to explore more diverse types of tasks.

\bibliography{anthology,custom}

\appendix

\section{Appendix}
\label{sec:appendix}

\subsection{Prompts for Different Methods} \label{sec:example_prompt_allmethods}

\begin{figure*}[t]
    \centering
    \includegraphics[width=0.88\textwidth]{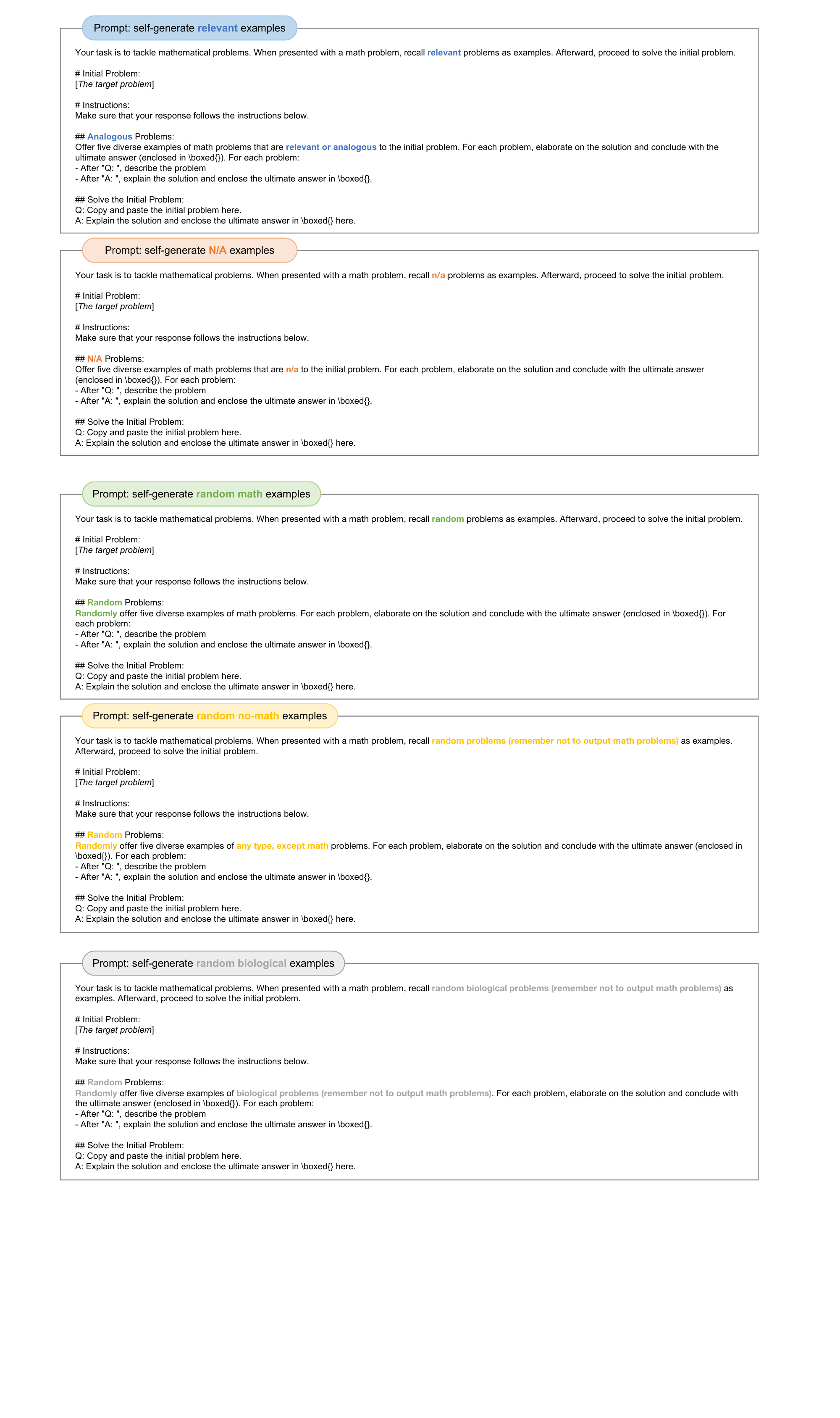}
    \caption{Prompts for different methods on GSM8K.}
    \label{fig:allprompt_gsm8k}
\end{figure*}

\begin{figure*}[t]
    \centering
    \includegraphics[width=0.88\textwidth]{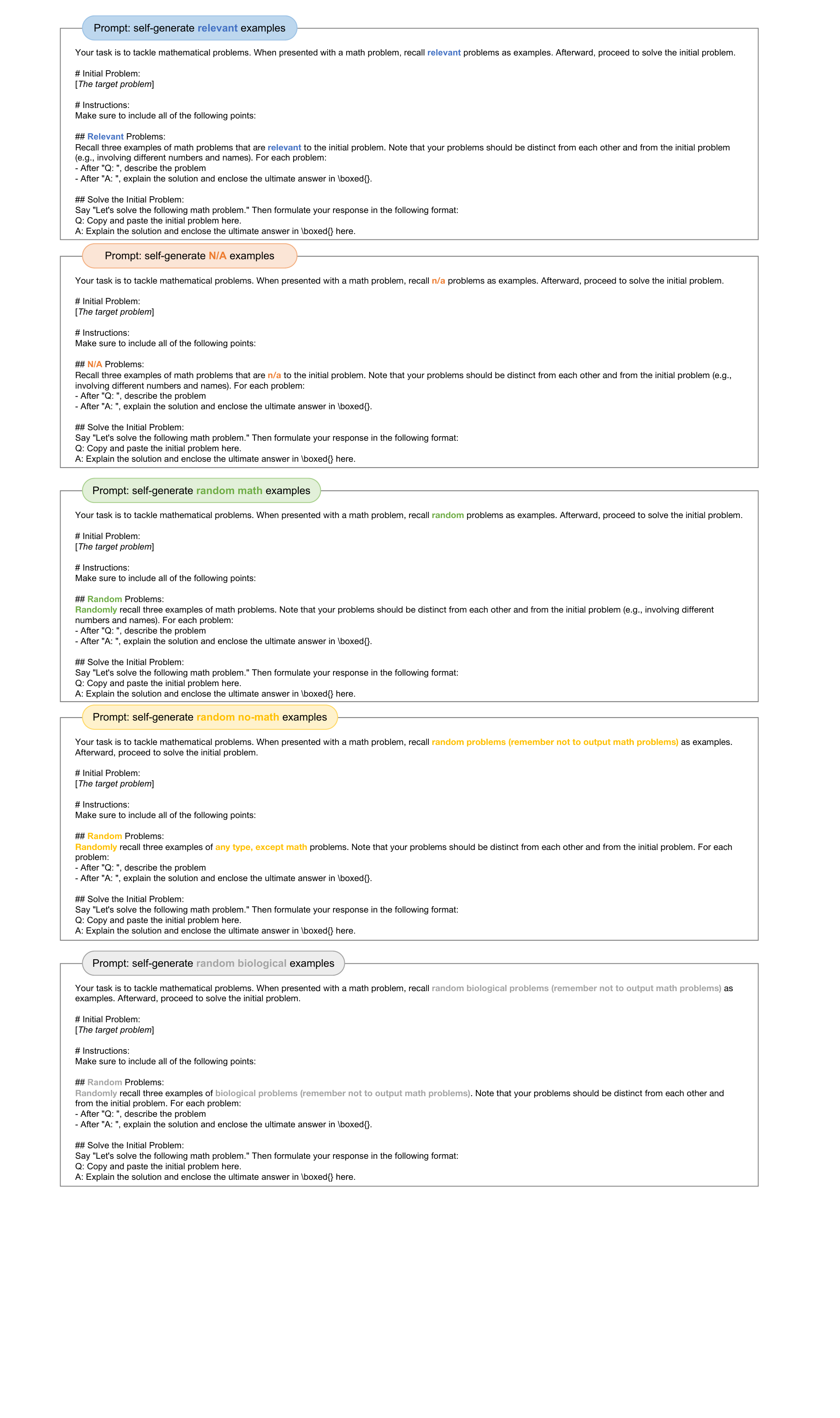}
    \caption{Prompts for different methods on MATH.}
    \label{fig:allprompt_math}
\end{figure*}

\begin{figure*}[t]
    \centering
    \includegraphics[width=0.88\textwidth]{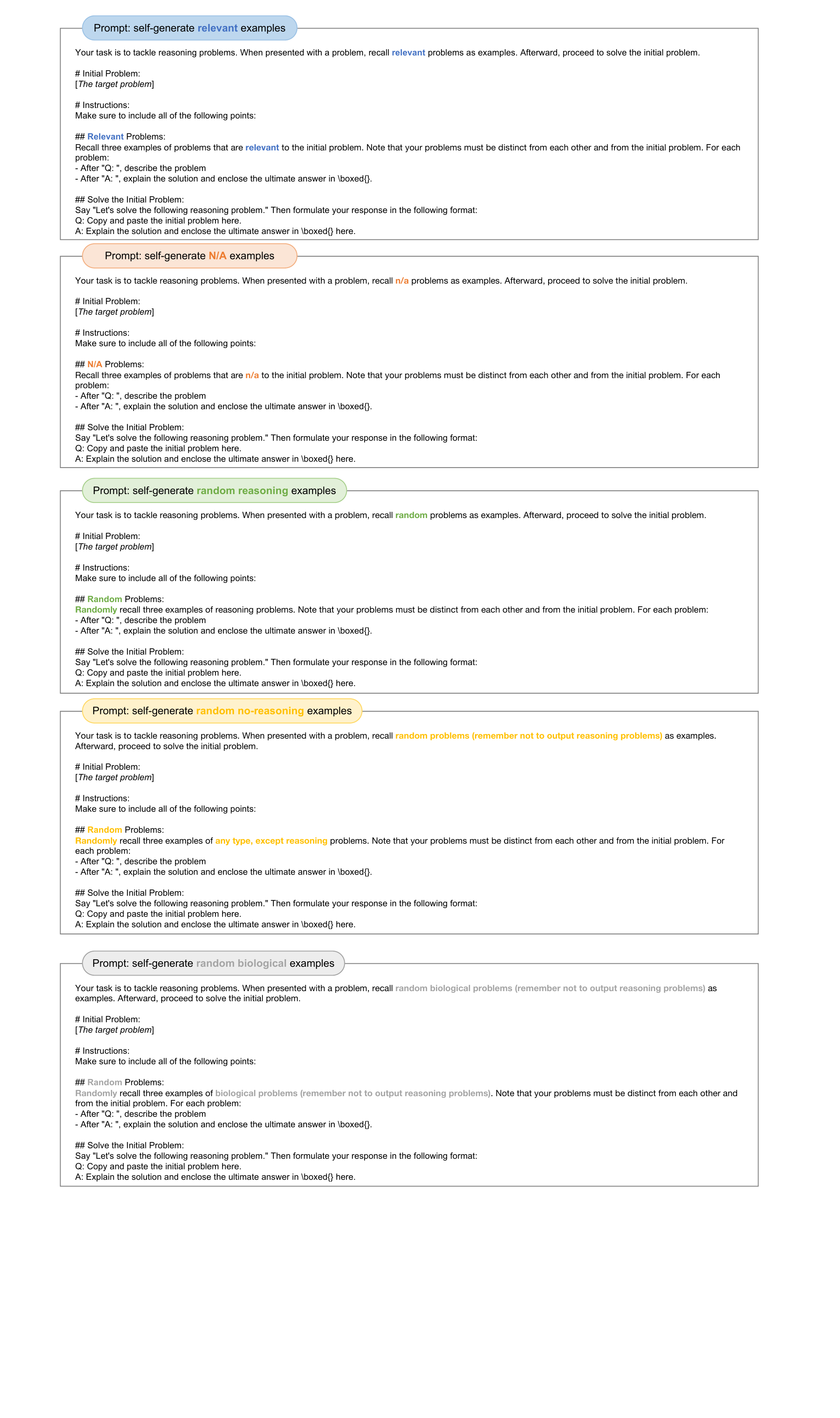}
    \caption{Prompts for different methods on BBH.}
    \label{fig:allprompt_bbh}
\end{figure*}

The prompts for different methods on all datasets are shown in \Cref{fig:allprompt_gsm8k} $\sim$ \Cref{fig:allprompt_bbh}.

\subsection{Detailed Results for Different Random Seeds} \label{sec:detailed_res_diff_seeds}

\begin{table*}[!t]
	\centering
	\scalebox{0.67}
	{
	\begin{tabular}{ll | ccccc | ccccc}
    \toprule
    \multicolumn{2}{c}{\multirow{2}*{\textbf{Seed}}} & \multicolumn{5}{c}{GSM8K}  & \multicolumn{5}{c}{MATH} \\
    \cmidrule{3-12}
    \multicolumn{2}{c}{} & Relevant & N/A & $\text{Random}_{\text{same}}$ & $\text{Random}_{\text{diff}}$ & $\text{Random}_{\text{bio}}$ & Relevant & N/A & $\text{Random}_{\text{same}}$ & $\text{Random}_{\text{diff}}$ & $\text{Random}_{\text{bio}}$  \\ 
    \midrule
    \multicolumn{2}{c}{42} & 71.8 & \textbf{76.6} & 73.2 & 74.0 & 74.0 & 37.4 & \textbf{42.2} & 41.6 &39.0  & 39.2 \\ 
    \midrule
    \multicolumn{2}{c}{100} & 71.2 & 75.2 & 75.2 & \textbf{75.8}  & 74.8 & 29.0 &	30.6 & \textbf{32.6} & 29.4	& 31.2 \\ 
    \midrule
    \multicolumn{2}{c}{1000} & 71.4	&74.8	&77.0	&\textbf{79.2}	&77.0 & 33.6	&\textbf{35.6}	&	34.6&	34.0&33.4 \\ 
    \midrule
    \multicolumn{2}{c}{Average} & $\text{71.5}_{\pm \text{0.3}}$	& $\text{75.5}_{\pm \text{0.8}}$	& $\text{75.1}_{\pm \text{1.5}}$	& $\textbf{76.3}_{\pm \text{2.1}}$	& $\text{75.3}_{\pm \text{1.2}}$ & 	$\text{33.3}_{\pm \text{3.4}}$ &	$\text{36.1}_{\pm \text{4.7}}$  &	$\textbf{36.3}_{\pm \text{3.8}}$ & $\text{34.1}_{\pm \text{3.9}}$	& $\text{34.6}_{\pm \text{3.3}}$ \\ 
    \bottomrule
    \end{tabular}
    } 
    \caption{Accuracy ($\%$) of all methods with different random seeds on two mathematical reasoning tasks.}
    \label{tab:detailed_results_all_seeds_math}
\end{table*}

\begin{table*}[t]    
    \centering
    \scalebox{0.88}{
    \begin{tabular}{
        llcccccc
        }
        \toprule
        \multicolumn{2}{l}{\textbf{Seed}} & \makecell{Temporal \\ sequences}  & \makecell{Logical deduction \\ five objects} & \makecell{Reasoning about \\ colored objects} & \makecell{Formal \\ fallacies} &  \makecell{Word \\ sorting} & Average \\
        \midrule
        \multirow{5}{*}{42}  & Relevant & \textbf{58.0} & \textbf{52.8} & 76.0 & 50.4 & \textbf{77.2} & \textbf{62.9}  \\
        & N/A & 56.4 & 44.8 & \textbf{77.6} & \textbf{54.0} & 76.8 & 61.9  \\
        & $\text{Random}_{\text{same}}$ & 52.4 & 48.8 & 74.8 & 51.6 & 72.8 & 60.1  \\
        & $\text{Random}_{\text{diff}}$ & 43.2 & 46.8 & 74.0 & 52.4 & 67.6 & 56.8  \\
        & $\text{Random}_{\text{bio}}$  & 56.8 & 52.0 & 74.0 & 52.0 & 76.4 & 62.2  \\
        \midrule
        \multirow{5}{*}{100}  & Relevant & \textbf{58.4} & \textbf{50.8} & \textbf{78.4} & 51.2 & 76.8 & \textbf{63.1}  \\
        & N/A & 55.2 & 46.0 & 74.8 & 52.8 & \textbf{79.2} & 61.6  \\
        & $\text{Random}_{\text{same}}$ & 50.8 & 48.4 & 73.6 & \textbf{53.2} & 75.2 & 60.2   \\
        & $\text{Random}_{\text{diff}}$ & 46.4 & 46.8 & 72.8 & 50.0 & 70.4 & 57.3  \\
        & $\text{Random}_{\text{bio}}$  & 58.0 & 48.4 & \textbf{78.4} & 51.2 & 73.6 & 61.9 \\
        \midrule
        \multirow{5}{*}{1000}  & Relevant & \textbf{63.6} & \textbf{50.0} & 75.6 & 52.0 & 76.8 & \textbf{63.6}  \\
        & N/A & 60.8 & 45.2 & 74.0 & \textbf{53.2} & \textbf{77.2} & 62.1  \\
        & $\text{Random}_{\text{same}}$ & 56.0 & 49.2 & 72.0 & 52.4 & 74.4 & 60.8   \\
        & $\text{Random}_{\text{diff}}$ & 43.2 & 40.8 & 70.4 & 51.2 & 69.6 & 55.0  \\
        & $\text{Random}_{\text{bio}}$  & 56.4 & 48.0 & \textbf{76.0} & 49.2 & 74.8 & 60.9  \\
        \midrule
        \multirow{5}{*}{Average}  & Relevant & $\textbf{60.0}_{\pm \text{2.6}}$ & $\textbf{51.2}_{\pm \text{1.2}}$ & $\textbf{76.7}_{\pm \text{1.2}}$ & $\text{51.2}_{\pm \text{0.7}}$ & $\text{76.9}_{\pm \text{0.2}}$ & $\textbf{63.2}_{\pm \text{0.3}}$ \\
        & N/A & $\text{57.5}_{\pm \text{2.4}}$ & $\text{45.3}_{\pm \text{0.5}}$ & $\text{75.5}_{\pm \text{1.5}}$ & $\textbf{53.3}_{\pm \text{0.5}}$ & $\textbf{77.7}_{\pm \text{1.0}}$ & $\text{61.9}_{\pm \text{0.2}}$  \\
        & $\text{Random}_{\text{same}}$ & $\text{53.1}_{\pm \text{2.1}}$ & $\text{48.8}_{\pm \text{0.3}}$ & $\text{73.5}_{\pm \text{1.1}}$ & $\text{52.4}_{\pm \text{0.6}}$ & $\text{74.1}_{\pm \text{1.0}}$ & $\text{60.4}_{\pm \text{0.3}}$ \\
        & $\text{Random}_{\text{diff}}$ & $\text{44.3}_{\pm \text{1.5}}$ & $\text{44.8}_{\pm \text{2.8}}$ & $\text{72.4}_{\pm \text{1.5}}$ & $\text{51.2}_{\pm \text{1.0}}$ & $\text{69.2}_{\pm \text{1.2}}$ & $\text{56.4}_{\pm \text{1.0}}$  \\
        & $\text{Random}_{\text{bio}}$  & $\text{57.1}_{\pm \text{0.7}}$ & $\text{49.5}_{\pm \text{1.8}}$ & $\text{76.1}_{\pm \text{1.8}}$ & $\text{50.8}_{\pm \text{1.2}}$ & $\text{74.9}_{\pm \text{1.1}}$ & $\text{61.7}_{\pm \text{0.6}}$  \\
        \bottomrule
    \end{tabular}
    }
    \caption{ Accuracy ($\%$) of all methods with different random seeds on BBH.
    }
    \label{tab:detailed_results_all_seeds_bbh}
\end{table*}

We report detailed results for different random seeds in \Cref{tab:detailed_results_all_seeds_math} $\sim$ \Cref{tab:detailed_results_all_seeds_bbh}.

\begin{table*}[t]    
    \centering
    \scalebox{0.86}{
    \begin{tabular}{lccccccc}
        \toprule
         \textbf{Method} & Relevant  & N/A & $\text{Random}_{\text{same}}$ & $\text{Random}_{\text{diff}}$ &  $\text{Random}_{\text{bio}}$ & $\text{ICL}_{\text{math}}$ &  $\text{ICL}_{\text{bio}}$ \\
        \midrule
        GPT-4o-mini &  90.7 & 91.9 &   92.6 & 92.3 & 93.2 & 94.2 & \textbf{94.5} \\
        \bottomrule
    \end{tabular}
    }
    \caption{Accuracy ($\%$) of different methods on GSM8K using GPT-4o-mini. Self-generated relevant examples always underperform irrelevant ones and both ICL variants perform better than other approaches.
    }
    \label{tab:diff_model_gsm8k_gpt4omini}
\end{table*}

\subsection{Results with GPT-4o-mini} \label{sec:results_gpt_4o_mini}

We conduct experiments with GPT-4o-mini on GSM8K and present the results in \Cref{tab:diff_model_gsm8k_gpt4omini}, verifying the generalizability of our findings to GPT-4o-mini.

\begin{table}[t]    
    \centering
    \scalebox{0.58}{
    \begin{tabular}{cccccccc}
        \toprule
         & Relevant  & N/A & $\text{Random}_{\text{same}}$ & $\text{Random}_{\text{diff}}$ &  $\text{Random}_{\text{bio}}$ & Oracle\\
        \midrule
        GSM8K & \textbf{0.50} & 0.16  & 0.28  & 0.19  & 0.08 & 0.62\\
        \bottomrule
    \end{tabular}
    }
   \caption{
\label{tab:decomposition_result}
Procedure (reasoning steps) relevance between self-generated examples and the query.
}
\end{table}

\subsection{Decomposition of Relevance} \label{sec:decomposition_similarity}

The relevance can be further separated into semantic relevance and procedure (reasoning steps) relevance. Our analysis in Q1-1 has demonstrated that semantic relevance does not matter. To investigate the importance of procedure relevance, we perform a similar analysis. Specifically, we compute the average cosine similarity between the rationales of the generated examples and the rationale of the query to quantitatively measure their relevance. The results on GSM8K are reported in \Cref{tab:decomposition_result}, which highlight that procedure relevance is not the key factor for analogical reasoning performance on mathematical reasoning tasks.

\subsection{Prompts and Outputs for Example Generation} \label{sec:detail_prompt_generating_problems}

\begin{figure*}[t]
    \centering
    \includegraphics[width=0.88\textwidth]{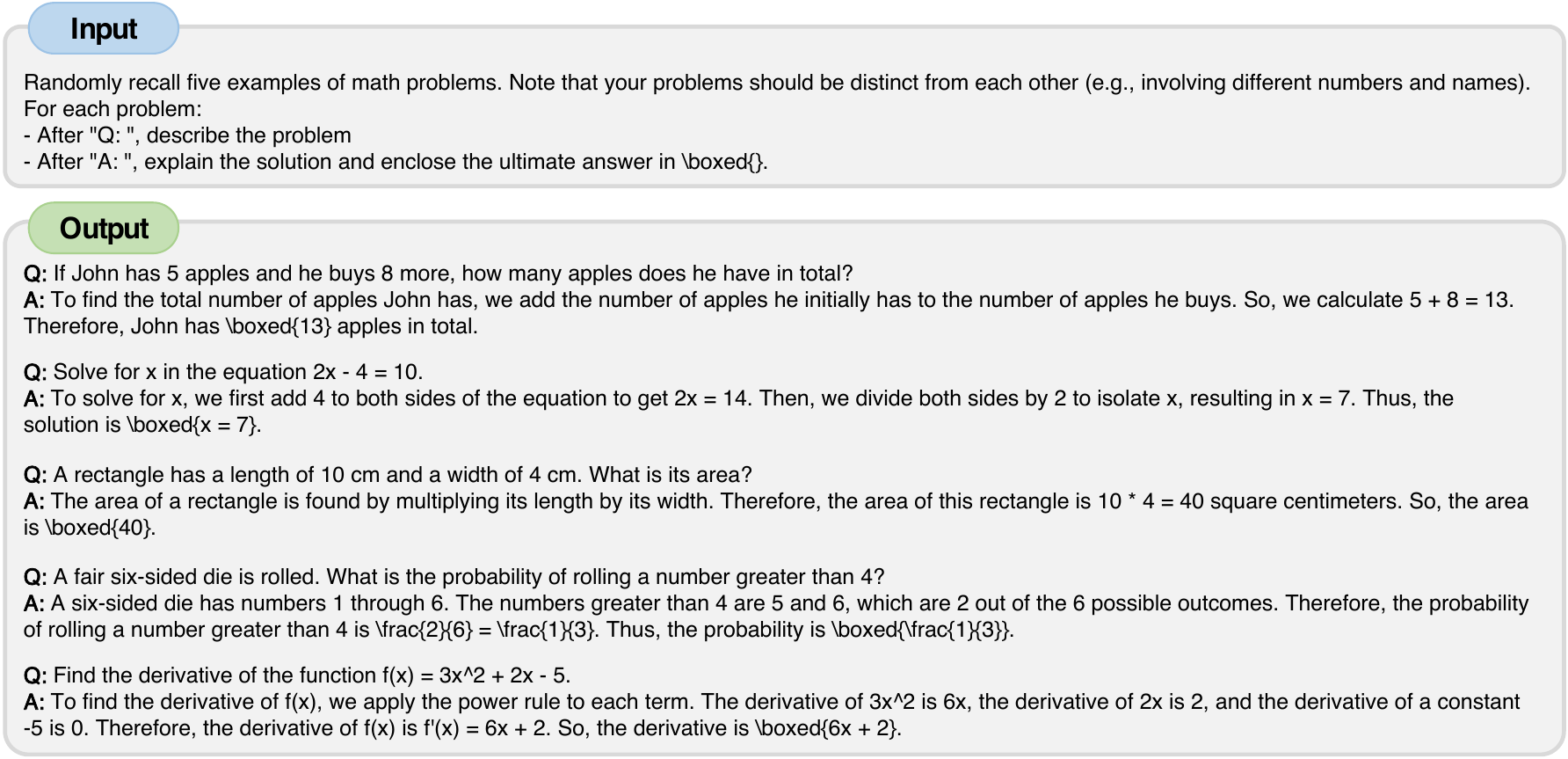}
    \caption{Prompts and outputs for generating math problems.}
    \label{fig:prompt_generating_math}
\end{figure*}

\begin{figure*}[t]
    \centering
    \includegraphics[width=0.88\textwidth]{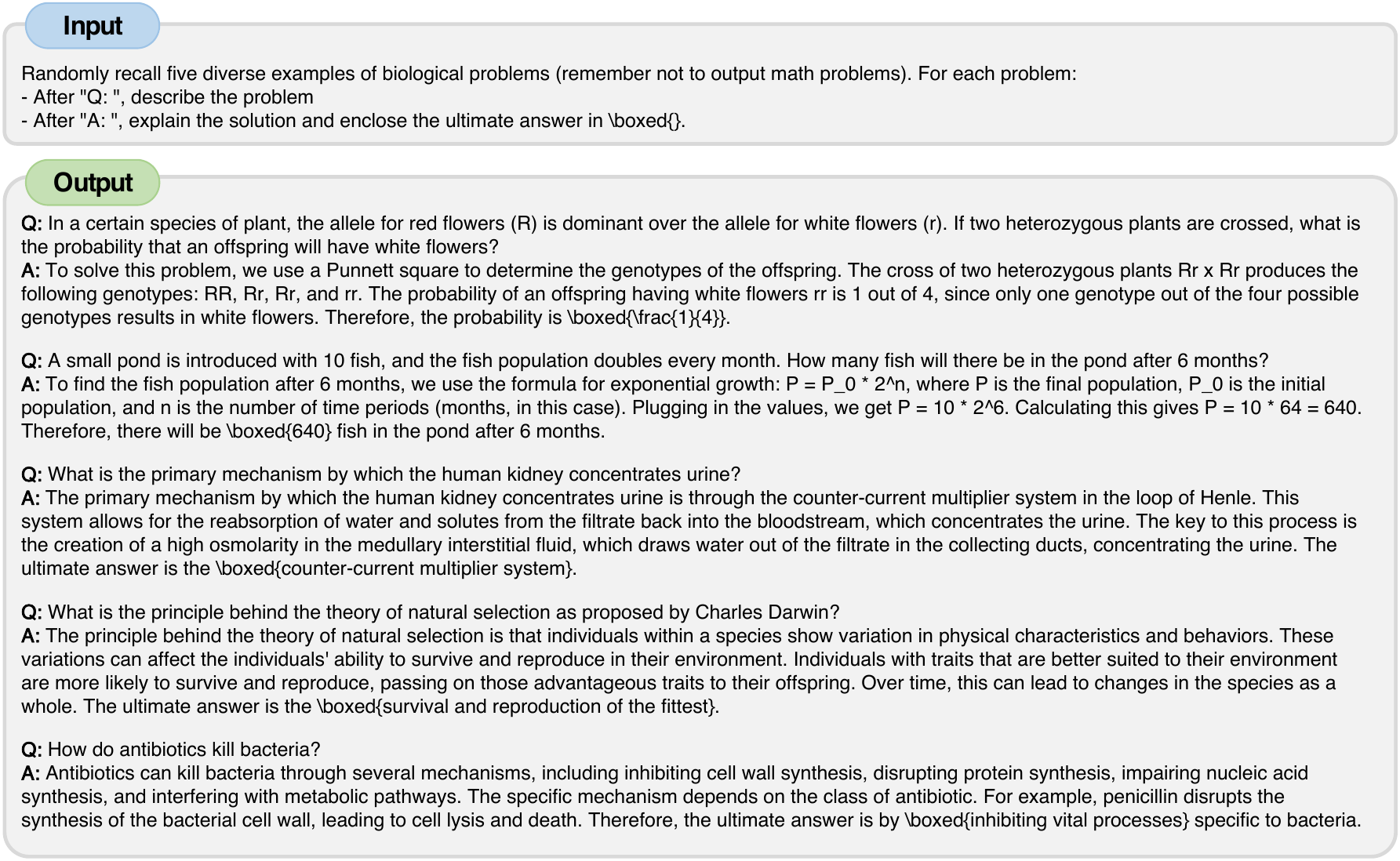}
    \caption{Prompts and outputs for generating biological problems.}
    \label{fig:prompt_generating_biological}
\end{figure*}

We show detailed prompts and outputs for randomly generating math and biological problems in \Cref{fig:prompt_generating_math} and \Cref{fig:prompt_generating_biological}, respectively.

\subsection{Guided Problem Generation} \label{sec:guided_problem_generation}

In addition to random problem generation in \Cref{sec:main_res_analogical}-Q2-1, we further investigate guided problem generation. Specifically, we randomly select 5 training samples to guide LLMs to self-generate relevant math problems. We then manually verify their correctness and use this fixed set of problems as ICL demonstrations for experiments. The performance of this approach (56.1) is slightly lower than that of $\text{ICL}_{\text{math}}$ (56.3), verifying that relevance is not the key influencing factor.

\begin{table}[t]    
    \centering
    \scalebox{0.68}{
    \begin{tabular}{lccccc}
        \toprule
         & Relevant  & N/A & $\text{Random}_{\text{same}}$ & $\text{Random}_{\text{diff}}$ &  $\text{Random}_{\text{bio}}$\\
        \midrule
        $\text{Prompt}_{1}$ & 71.2 & 74.9  & 75.3 & \textbf{75.9} &  74.3 \\
        $\text{Prompt}_{2}$ & 72.0 & 75.2  & 74.7 & \textbf{76.2} & 75.5  \\
        \bottomrule
    \end{tabular}
    }
   \caption{
\label{tab:new_prompt_res}
Accuracy ($\%$) of different methods with two new prompts.
}
\end{table}

\begin{table}[t]    
    \centering
    \scalebox{0.65}{
    \begin{tabular}{lccccc}
        \toprule
          \textbf{Number} & Relevant  & N/A & $\text{Random}_{\text{same}}$ & $\text{Random}_{\text{diff}}$ &  $\text{Random}_{\text{bio}}$\\
        \midrule
        3 & 73.1 & \textbf{77.3}  & 75.0 & 75.3 & 75.5  \\
        5 & 71.5 & 75.5 & 75.1 & \textbf{76.3} & 75.3  \\
        \bottomrule
    \end{tabular}
    }
   \caption{
\label{tab:diff_num_examples}
Accuracy ($\%$) of all methods with different numbers of demonstrations.
}
\end{table}

\begin{table}[t]    
    \centering
    \scalebox{0.86}{
    \begin{tabular}{cccc}
        \toprule
         $\text{ICL}_{\text{math}}$ & $\text{ICL}_{\text{math}}^{\text{wrong}}$ & $\text{ICL}_{\text{bio}}$ &   $\text{ICL}_{\text{bio}}^{\text{wrong}}$\\
        \midrule
        56.3  & 50.9  & 56.4 & 51.3\\
        \bottomrule
    \end{tabular}
    }
   \caption{
\label{tab:more_analysis_result}
Comparison between different ICL variants.
}
\end{table}

\begin{table}[t]
\centering
    \scalebox{0.86}{
    \begin{tabular}{lccc}
    \toprule
    \multirow{2}{*}{\textbf{Method}} & \multicolumn{3}{c}{\textbf{Task}} \\
    \cmidrule(lr){2-4}
    & GSM8K & MATH & Average  \\
    \midrule
    $\text{ICL}_{\text{math}}$ & \textbf{75.7} & \textbf{36.8} & \textbf{56.3} \\
    $\text{ICL}_{\text{math\_repeat}}$ & 73.8 & 36.2 & 55.0 \\
    \bottomrule
    \end{tabular}
    }
\caption{
\label{tab:comparison_repeating_problem}
Comparison of two ICL variants on the GSM8K and MATH datasets.
}
\end{table}

\subsection{Robustness to Prompt Format} \label{sec:robust_prompt}

To verify the robustness of different methods to prompt format, we experiment with two new prompts paraphrased from the original one by GPT-4 and present the results on GSM8K in \Cref{tab:new_prompt_res}. We also observe better performance with irrelevant examples than relevant ones, showing the robustness.

\subsection{Different Numbers of Demonstrations} \label{sec:diff_num_demonstration}

While we mainly follow the setting in \citet{yasunaga2024large} to ask the LLM to generate $k = 5$ examples for GSM8K, we further investigate the effect of the number of demonstrations. Specifically, we conduct controlled experiments with $k = 3$ and report the results in \Cref{tab:diff_num_examples}. We can observe that irrelevant examples consistently outperform relevant ones across different numbers of demonstrations, emphasizing their effectiveness.

\subsection{More Analysis on $\text{ICL}_{\text{math}}$ and $\text{ICL}_{\text{bio}}$} \label{sec:more_analysis_icl}

Our designed method $\text{ICL}_{\text{math}}$ generates \emph{correct and relevant} examples, and $\text{ICL}_{\text{bio}}$ generates \emph{correct and irrelevant} examples. From the results in \Cref{tab:comparison_icl}, we can see that $\text{ICL}_{\text{math}}$ and $\text{ICL}_{\text{bio}}$ achieve similar average performance, demonstrating that relevance does not matter.

We further change the correct answers of the demonstrations in $\text{ICL}_{\text{math}}$ and $\text{ICL}_{\text{bio}}$ to random answers, obtaining $\text{ICL}_{\text{math}}^{\text{wrong}}$ and $\text{ICL}_{\text{bio}}^{\text{wrong}}$. Obviously, $\text{ICL}_{\text{math}}^{\text{wrong}}$ generates \emph{incorrect and relevant} examples, and $\text{ICL}_{\text{bio}}^{\text{wrong}}$ generates \emph{incorrect and irrelevant} examples. The comparison between these four methods in \Cref{tab:more_analysis_result} further supports our claim that the key factor influencing the performance on mathematical reasoning is the accuracy of self-generated examples rather than their relevance.

\subsection{Repeating Problems} \label{sec:repeating_problems}

While generating a few accurate problems as ICL demonstrations can achieve better performance, a bolder idea might be to generate one problem and repeat it multiple times as few-shot demonstrations for ICL. 
To investigate this, we randomly select a generated math problem and repeat it to perform ICL, denoted by  $\text{ICL}_{\text{math\_repeat}}$. From the results shown in \Cref{tab:comparison_repeating_problem}, we can see that $\text{ICL}_{\text{math\_repeat}}$ consistently performs worse than $\text{ICL}_{\text{math}}$ on both datasets, indicating that the diversity of generated problems also matters.

\begin{table}[t]    
    \centering
    \scalebox{0.60}{
    \begin{tabular}{ccccc}
        \toprule
         Relevant  & N/A & $\text{Random}_{\text{same}}$ & Similar and Correct &  Different and Correct \\
        \midrule
         74.8 & 77.3  &  78.4  & 80.3 & \textbf{80.6}  \\
        \bottomrule
    \end{tabular}
    }
   \caption{
\label{tab:explicit_semantic_control}
Accuracy ($\%$) of different methods on GSM8K using Llama-3.1-8B-Instruct.
}
\end{table}

\subsection{Explicit Semantic Control} \label{sec:explicit_semantic}

We explore explicitly controlling the semantics of generated examples (including both problems and reasoning paths) on GSM8K using Llama-3.1-8B-Instruct. Specifically, we investigate the following two approaches: \Ni prompting the model to generate \emph{semantically similar and correct} examples, and \Nii prompting the model to generate \emph{semantically different and correct} examples. The results reported in \Cref{tab:explicit_semantic_control} further verify the correctness of our conclusions.

\end{document}